\documentclass[journal]{IEEEtran}
\usepackage{lineno}

\usepackage{url}
\usepackage{xcolor}
\usepackage{color}
\usepackage{enumerate}

\usepackage{epsfig}
\usepackage{graphicx}
\usepackage{amsmath}
\usepackage{times}
\usepackage{array}
\usepackage{subfigure}
\usepackage{multirow}
\usepackage{booktabs}
\usepackage{threeparttable}
\newcommand{\PreserveBackslash}[1]{\let\temp=\\#1\let\\=\temp}
\newcolumntype{C}[1]{>{\PreserveBackslash\centering}p{#1}}
\newcolumntype{R}[1]{>{\PreserveBackslash\raggedleft}p{#1}}
\newcolumntype{L}[1]{>{\PreserveBackslash\raggedright}p{#1}}

\ifCLASSINFOpdf
\else
\fi
\begin{document}
%
\title{Face Presentation Attack Detection in Learned \\Color-liked Space}
%
%
%

\author{Lei Li,
        Zhaoqiang Xia$^*$,
        Xiaoyue Jiang,
        Fabio Roli,
        and Xiaoyi Feng$^*$
\thanks{L. Li, Z. Xia, X. Jiang, and X. Feng are with Northwestern Polytechnical University, Xi'an 710129, China. e-mail: (lilei\_npu@mail.nwpu.edu.cn; zxia@nwpu.edu.cn; xjiang@nwpu.edu.cn; fengxiao@nwpu.edu.cn).}
\thanks{F. Roli is with the University of Cagliari, Italy. e-mail: (roli@diee.unica.it).}
\thanks{$*$ Corresponding authors.}
}

\maketitle

\begin{abstract}
  Face presentation attack detection (PAD) has become a thorny problem for biometric systems and numerous countermeasures have been proposed to address it. However, majority of them directly extract feature descriptors and distinguish fake faces from the real ones in existing color spaces (e.g. RGB, HSV and YCbCr). Unfortunately, it is unknown for us which color space is the best or how to combine different spaces together. To make matters worse, the real and fake faces are overlapped in existing color spaces. So, in this paper, a learned distinguishable color-liked space is generated to deal with the problem of face PAD. More specifically, we present an end-to-end deep learning network that can map existing color spaces to a new learned color-liked space. Inspired by the generator of generative adversarial network (GAN), the proposed network consists of a space generator and a feature extractor. When training the color-liked space, a new triplet combination mechanism of points-to-center is explored to maximize interclass distance and minimize intraclass distance, and also keep a safe margin between the real and presented fake faces. Extensive experiments on two standard face PAD databases, i.e., Relay-Attack and OULU-NPU, indicate that our proposed color-liked space analysis based countermeasure significantly outperforms the state-of-the-art methods and show excellent generalization capability.
\end{abstract}


%
\IEEEpeerreviewmaketitle

\section{Introduction}
%
%
%
%

\IEEEPARstart{A}{s} one of the most natural clues for identifying individuals, faces have been used as the biometric trait in many biometric recognition systems \cite{Akhtar2012Evaluation,Wen2015Face}, especially since face recognition technology has reached a satisfactory level of performance. Nowadays, large scale face recognition systems can be easily to be found, such as the Unique Identification Authority of India (UIDAI), which offers identity to all India residents \cite{Li2018Face2}. Although these biometric systems have achieved high accuracy on recognizing customer faces, fake faces still can easily fool or bypass them \cite{Li2014Understanding,Omar2015Evaluating}. With the popularity of Internet communication and social media, people's biological information can be rapidly proliferated leading to the criminals can easily access to face pictures or videos. Actually, once a face image is shared in the Internet, there is no further control over that image. For instance, Olaye \cite{Olaye2018Biometrics} has reported that a mobile app called FindFace allows uploading a picture to access that person's social network data including good quality images that can be used for face PAD. To make matters even worse, it is not difficult for the criminals to use the biological information to launch a presentation attack. Therefore, considering this urgent security situation, an effective and reliable face PAD method must be developed for circumventing and detecting such threats.

\begin{figure}[!t]
 \centering
 \includegraphics[width=0.48\textwidth,angle=0]{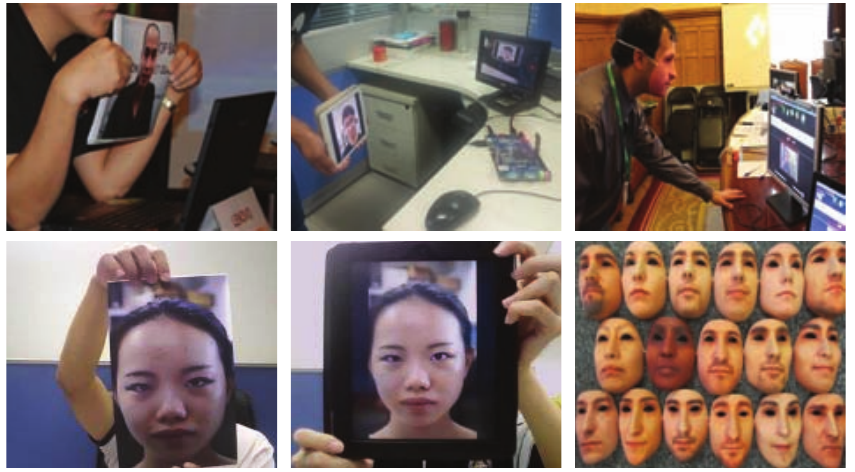}
 \caption{Samples of face presentation attacks. From left to right: printed face photos, displayed face images or replayed videos, and 3D masks.}\label{fig:differentAttacks}
\end{figure}

According to different kinds of fake faces, there are four types of presentation attacks that can be considered: (i) printed face photos, (ii) displayed face images, (iii) replayed videos and (iv) 3D masks. In printed face photo attacks, the face photos are printed on paper and placed in front of the camera. Sometimes in order to produce aliveness signals such as eye blinking to confuse face recognition system, eye area of the photo will be cut off and replaced with real eye blinking. In both displayed image and replayed video attacks, the attacker uses a digital screen to show face images or videos. Compared with printed photo and displayed image attacks, replayed video attacks can exhibit motion and liveness information and are more challenging on common cameras. 3D mask scenario refers to utilizing 3D printing technology \cite{manjani2017detecting} and virtual reality \cite{xu2016virtual} to fabricate a face mask and presenting the mask for presentation attack. However, 3D attacks are much more expensive to launch compared with traditional printed photo, displayed image and replayed video attacks \cite{Li2018Unsupervised}. Fig. \ref{fig:differentAttacks} shows an example of different face presentation attacks.

\begin{figure}[t]
 \centering
 \includegraphics[width=0.48\textwidth,angle=0]{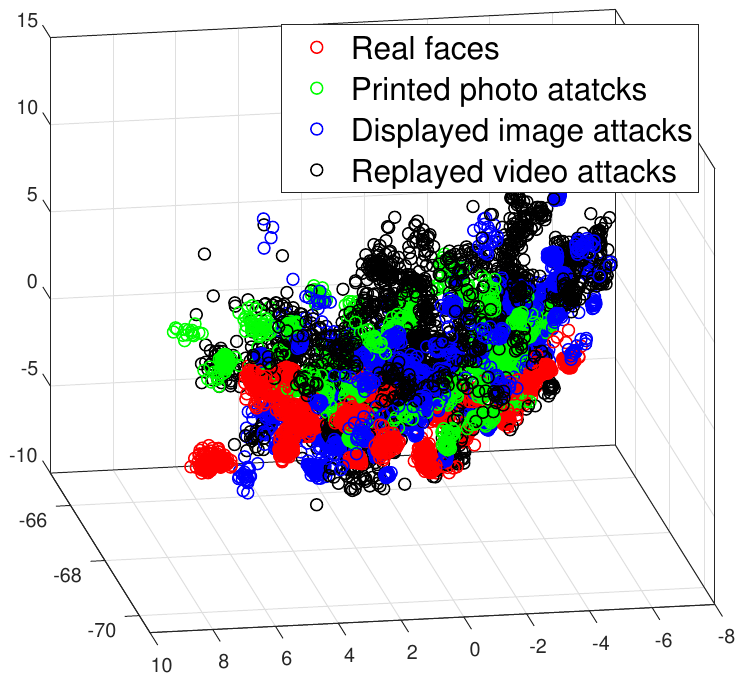}
 \caption{The distribution of some real and fake faces of Replay-Attack database \cite{Chingovska2012On} in RGB color space. The red circles are the real faces. The green, blue and black circles are fake faces. In this figure, we extract features from the pre-trained VGG-19 model \cite{simonyan2014very} and visualize them in a 3D space.}
 \label{fig:RGB_distribution}
\end{figure}

In the last decade, many face PAD methods have been proposed to detect fake faces \cite{Li2018Face2, manjani2017detecting, xu2016virtual, Li2018Unsupervised, Chingovska2012On,Erdogmus2013Spoofing,Maatta2011Face,Boulkenafet2015Face,Pan2007Eyeblink,Li2009An,Zhang2012A,Tan2010Face,Li2017Face,Pavlidis2000The,Zhang2011Face}. However, most methods directly extracted features and analyzed the differences between real and fake faces in existing color spaces (e.g. RGB, HSV and YCbCr). For instance, \cite{Chingovska2012On,Maatta2011Face} distinguished real faces with the fake ones in gray-scaled space and \cite{Li2018Unsupervised,Boulkenafet2015Face} detected presented fake faces in RGB color space. Although these methods can achieve good detection performance, the real faces and fake faces are overlapped in original color spaces as shown in Fig. \ref{fig:RGB_distribution}. Especially with the popularity of high-resolution screen and high-definition printer, the fake faces are getting closer to the real faces leading to detection task becomes more difficult. Therefore, we propose an end-to-end deep learning network to generate a new color-liked space, where the real and fake faces can be separated as much as possible.

Inspired by generative adversarial network (GAN) \cite{Goodfellow2014Generative}, a color-liked space generator is constructed to map existing color spaces. Then, a feature extractor is designed to obtain features from the learned color-liked space. As aforementioned, the goal of our proposed network is to separate the real and fakes as much as possible. Therefore, we introduce a novel points-to-center triplet mechanism to train the generator. The flowchart is illustrated in Fig. \ref{fig:flowchart}. Finally, the extracted features are fed into a Support Vector Machine (SVM) \cite{Cortes1995Support} classifier to detect face presentation attack. We train and test our proposed method on two public available databases: Replay-Attack \cite{Chingovska2012On} and OULU-NPU \cite{Boulkenafet2017OULU}. The experimental results demonstrate the effectiveness and excellent generalization capabilities of the proposed method in various fake face detection compared to the state-of-the-art approaches.

Among the significant contributions of this present paper, we can cite:

\begin{enumerate}
\item Different from most previous works that detect presentation attack in existing color spaces, we propose a novel and appealing approach based on color-liked space analysis. By maximizing interclass distance and minimizing intraclass distance, the learned color-liked space can separate fake faces and genuine ones as much as possible instead of overlapping each other.
\item We develop a new points-to-center triplet training mechanism. Unlike traditional mechanism, the proposed points-to-set mechanism can guarantee a stable decline in triplet loss.
\item Extensive experimental analysis is conducted on the two latest and challenging face PAD databases using their pre-defined publicly well-defined experimental evaluation protocols ensuring the reproducibility of the results and a fair comparison with the state-of-the-art methods. Furthermore, in our cross-database evaluation, our generated color-liked space shows promising generalization capabilities.
\end{enumerate}

The remainder of the paper is organized as follows: Section II reviews the existing state-of-the-art methods of face PAD and briefly provides the development of triplet network. Section III describes our proposed color-liked space analysis based method. Section IV provides the details of experimental setup and Section V discusses the obtained results. Finally, in Section VI, we conclude the paper and discuss some directions for future research.

\section{Related Work}
\subsection{State-of-the-art of Face PAD}
In the past few years, face PAD has received great attention and many detection approaches have been developed \cite{Chingovska2012On,Erdogmus2013Spoofing,Maatta2011Face,Boulkenafet2015Face,Pan2007Eyeblink,Li2009An,Zhang2012A,Tan2010Face,Li2017Face,Pavlidis2000The,Zhang2011Face}. Based on different clues, these countermeasures can be further categorized into texture analysis \cite{Chingovska2012On,Maatta2011Face,Boulkenafet2015Face}, motion analysis \cite{Pan2007Eyeblink,Li2009An,Anjos2011Counter}, image quality analysis \cite{Zhang2012A,Tan2010Face,Li2017Face,Pavlidis2000The,Zhang2011Face}, and hardware based methods \cite{Erdogmus2013Spoofing,Pavlidis2000The,Zhang2011Face,Kim2014Face,Ji2016LFHOG,Sepas2017Light}.

\subsubsection{Texture analysis based methods}
Due to the limitations of printers and display devices, there are substantial differences in color distribution and edge texture between real and fake faces. Therefore, multi-scale local binary pattern (LBP) features \cite{Maatta2011Face} were extracted to describe edge texture differences. In another work, Chingovska \textit{et al.} \cite{Chingovska2012On} used different kinds of LBP features to improve detection accuracy. To capture the color differences in luminance and chrominance, Boulkenafet \textit{et al.} \cite{Boulkenafet2015Face,Boulkenafet2016FaceTIFS} proposed a method by computing LBP features from different color spaces and concatenating all LBP into a single feature vector. Apart from \cite{Boulkenafet2015Face}, Boulkenafet \textit{et al.} \cite{Boulkenafet2016Scale} constructed a Gaussian pyramid and extracted multi-scale textural features from it. In this work, the method shows superior detection performance in inter-database test. Instead of using LBP features, Akshay \textit{et al.} \cite{Agarwal2016Face} extracted Haralick \cite{Haralick1973Textural} texture features for face PAD. With the recent advances of deep learning in computer vision \cite{Krizhevsky2012ImageNet,Xia2017Deep}, deep textures have also been applied in face PAD. Yang \textit{et al.} \cite{Yang2014Learn} proposed an end-to-end convolutional neural network (CNN) model for face PAD. Instead of using fully-connected layers, Li \textit{at al.} \cite{Li2017An,Li2018Face} extracted hand-crafted features from convolutional feature maps and showed some promising performances. For capturing texture variations, Xu \textit{et al.} \cite{Xu2016Learning} proposed a long short memory network (LSTM) and Li \textit{et al.} \cite{Li2018Learning} proposed a 3D CNN to detect face presentation attacks, respectively. In another work, Li \textit{at al.} \cite{Li2018Unsupervised} combined hand-crafted features with deep learning features and learned an embedding function that can measure the distribution similarity of different face images. More recently, a LBP network \cite{Li2018LBPnetwork} has been designed for face PAD which simulates the idea of basic LBP. These texture analysis based approaches have good detection performances when the artificial traces are obvious, such as rough face texture. However, with the popularity of high-definition screens, their detection performances tend to decrease drastically.

\subsubsection{Motion analysis based methods}
Apart from texture analysis, motion also plays an important role in face PAD. For instance, based on the fact that involuntary eyes blinking often occurs in the interval of 2 to 4 seconds \cite{karson1983spontaneous}, an undirected conditional random field framework \cite{Pan2007Eyeblink} was proposed to detect printed photo attacks. Compared with the fixed background of real face, the background of fake face is dithering. Therefore, Anjos \textit{et al.} \cite{Anjos2011Counter} detected fake face by analyzing motion correlation coefficient between face region and background. Moreover, facial motion variations were also explored for face PAD. In \cite{Pereira2012LBP-TOP} and \cite{Phan2016FACE}, LBP-TOP \cite{Zhao2007Dynamic} and LDP-TOP \cite{Zhang2010Local} features were extracted to describe these variations, respectively. In another work, Santosh \textit{et al.} \cite{Tirunagari2015Detection} used dynamic mode decomposition (DMD) to capture the dynamics of movements. Instead of analyzing facial motion, some works have tried to depict the media and its movements (for example the paper, mobile phone and laptop). Bao \textit{et al.} \cite{Bao2009A} addressed 2D face PAD by detecting planar media. Tan \textit{et al.} \cite{Li2009An} used Difference-of-Gaussians (DoG) to extract the differences in motion deformation patterns between real and fake faces. Since there are no aliveness signals in 3D mask attacks, Li \textit{et al.} \cite{Li2017Generalized} proposed an 3D mask PAD method by estimating pulse from face videos. Although motion analysis based methods are effective to static image attacks and 3D mask attacks, these methods can still be easily fooled by replayed video attacks. Therefore, it is necessary to request the subject to perform specific movements \cite{Pan2011Monocular,Smith2015Face}.

\subsubsection{Image quality analysis based methods}
Methods based on image quality analysis exploit the fact that the characteristics of reproduced face images or videos typically present lower quality and loss of detail and sharpness. Therefore, some methods took advantage of high frequency components of face image to recognize fake faces. In \cite{Tan2010Face} and \cite{Zhang2012A}, DoG filters were used to extract high frequency information. In another work, Li \textit{et al.} \cite{Li2017Face} dealt with the problem of face PAD based on a learned high frequency feature mapping function. Instead of high frequency information, Pinto \textit{et al.} \cite{Pinto2015Face} minimized possible negative impacts on low-level features and extracted them to describe the missing detail and sharpness in fake faces. Wen \textit{et al.} \cite{Wen2015Face} extracted four different kinds of features (i.e. specular reflection, blurriness, chromatic moment and color diversity) to catch the light reflection differences between real and fake faces, which are caused by the medium of liquid-crystal display (LCD) screen. Galbally \textit{et al.} \cite{Galbally2014Image} extracted different kinds of image quality assessment features to describe the quality of real and fake faces. In \cite{Feng2016Integration}, several different non-reference image-quality features and dense optical flow features were fused to capture the fake faces that have low quality. Such image quality analysis based approaches are expected to work well for low-resolution printed photo attacks or when using crude face masks, but are likely to fail for high quality displayed images, replayed videos or 3D masks.

\subsubsection{Hardware based methods}
In addition to analyzing face images and videos, additional or non-conventional hardwares, e.g., multi-spectral, depth and light-field cameras, have been applied to acquire useful information about the reflectance properties and the shape of the observed faces. For instance, based on the fact that there is no depth information in 2D presentation attacks, Erdogmus \textit{et al.} \cite{Erdogmus2013Spoofing} proposed a method by detecting a planar surface such as a video display or a printed paper (not bent). Pavlidis \textit{et al.} \cite{Pavlidis2000The} and Zhang \textit{et al.} \cite{Zhang2011Face} acquired the reflectance maps by computing the upper-band of near-infrared (NIR) spectrum and combining two photodiodes respectively, instead of using intrinsic image decomposition algorithm like \cite{Wen2015Face}. More recently, light-field cameras allow exploiting disparity and depth information from a single capture. Therefore, these kinds of cameras have also been introduced into face PAD task \cite{Kim2014Face,Ji2016LFHOG,Sepas2017Light}. Even though hardware-based methods can achieve good performances for replayed video attacks, some of them might present operation restrictions in certain conditions. For instance, the sunlight can cause severe perturbations for NIR and depth sensors; wearable 3D masks are obviously challenging for those methods relying on depth data.

\begin{figure*}[!t]
 \centering
 \includegraphics[width=0.85\textwidth,angle=0]{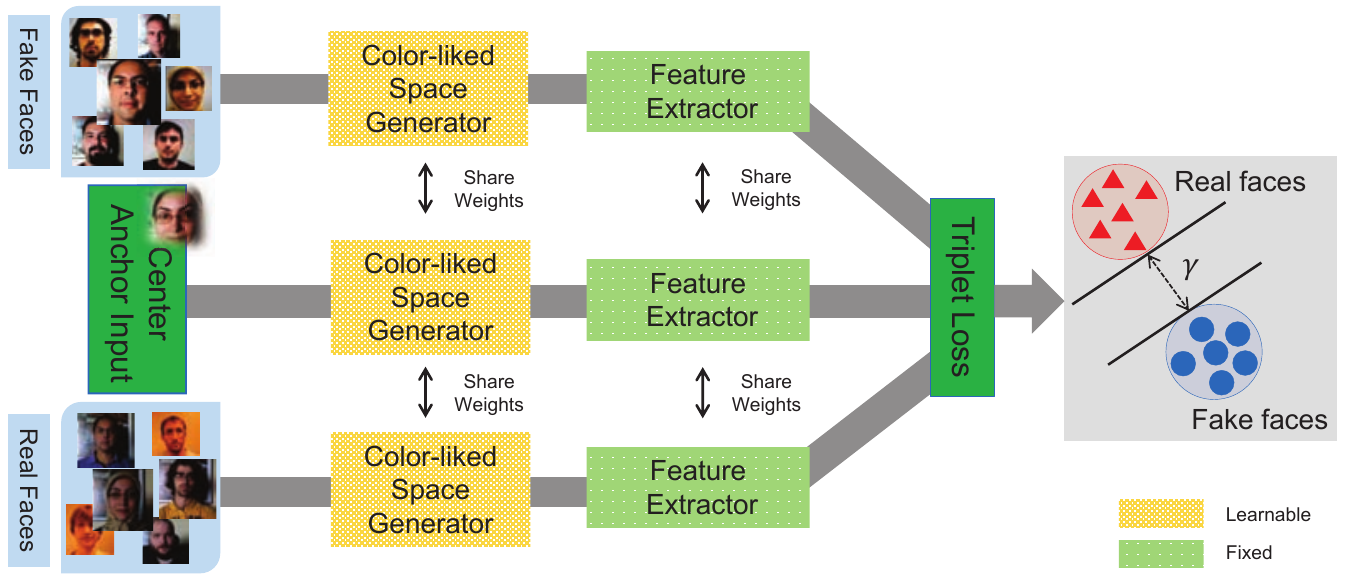}
 \caption{The flowchart of proposed face PAD method in learned color-liked space. First, the color-liked space generator is used to learn a new color space, which has three channels as existing RGB space. Then, a feature extractor module and triplet loss are invoked to maximize interclass distance and minimize intraclass distance.}
 \label{fig:flowchart}
\end{figure*}

\subsection{Triplet Network}
Triplet network \cite{Elad2014Deep,Florian2015FaceNet} is extended from siamese network \cite{bromley1994signature}, which aims to find a mapping function from original feature space to new distance space where the same class samples are similar than those from different ones. While ensuring maximizing interclass distances and minimizing intraclass distances, the triplet network can also maintain a secure margin between different classes compared to siamese network. In the past few years, many computer version and multimedia analysis tasks such as face verification and person re-identification (Re-ID) have explored the effectiveness of triplet network. For instance, Ding \textit{et al.} \cite{Ding2015Deep} used the triplet loss to learn a deep neural network for person Re-ID. Wang \textit{et al.} \cite{Wang2016Joint} addressed person Re-ID problem by presenting a unified siamese and triplet deep architecture which can jointly extract single-image and cross-image feature representations. In \cite{Swami2016Triplet}, Swami \textit{et al.} proposed a triplet network for face verification and got promising results. Wang \textit{et al.} \cite{Wang2015Unsupervised} designed a siamese-triplet convolutional neural network with a ranking loss function to learn visual representations from unlabeled videos. As far as we know, although triplet network can achieve better performance, there is no relevant work to introduce it into face PAD.

\section{Proposed Face PAD Method in Learned Color-Liked Space}
In this section, we present the pipeline of our proposed face PAD method that extracts features from a learned color-liked space. The proposed method consists of three parts, i.e., color-liked space generator, feature extractor and triplet loss. The overall process pipeline is shown in Fig. \ref{fig:flowchart}. Firstly, the captured RGB face image is fed into color-liked space generator, which can map existing color space into a new space. Then, the learned color-liked face image, real face images and fake face images are fed into the weights shared feature extractors to get deep features. Finally, the points-to-center triplet training mechanism is used to maximize interclass distance and minimize intraclass distance, and also keep a safe margin between different face classes. The following paragraphs explain the parts in more detail, respectively.

\subsection{Color-liked Space Generator}
\begin{figure*}[!t]
 \centering
 \includegraphics[width=0.8\textwidth,angle=0]{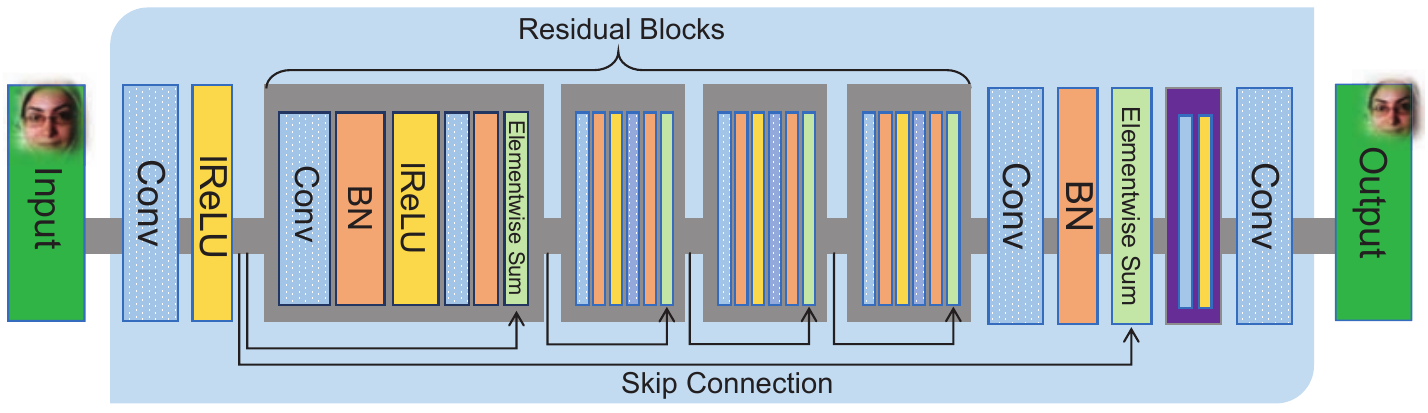}
 \caption{The architecture of proposed color-liked space generator. The generator consists of a series of concatenated residual blocks and the input of each residual block is skip connected to its outputs.}
 \label{fig:generator}
\end{figure*}

The architecture of color-liked space generator is shown in Fig. \ref{fig:generator}. In order to map existing color spaces and use pre-trained deep learning model to extract features, the convolutional feature maps with three channels are outputted by color-liked space generator. More specifically, given a color video frame $f(x,y)$, a convolutional layer with filter size of $3\times 3\times 3\times 64$ and a leaky rectified linear unit (lReLU) layer are firstly used to process $f(x,y)$. Then, these feature maps are passed through a series of concatenated residual blocks. In each residual block, the input is skip connected to its outputs as He \textit{et al.} \cite{He2016deep}. Finally, the resulting feature maps are passed through a convolutional layer with filter size of $3\times 3\times 64\times 3$ and the generated color-liked space of last layer is denoted as $f^{'}(x,y)=G(f(x,y),W_{1})$, where $G(\cdot)$ denotes the mapping function of the color-liked space generator and $W_1$ represents the generator parameters. It is noted that the size of all convolutional filters in the generator is $3\times 3$ allowing for deeper models at a low number of parameters \cite{Sajjadi2017enhancenet}. Moreover, the batch normalization (BN) layer \cite{Ioffe2015Batch} is introduced after convolutional layer. The hierarchic link of color-liked space generator is summarized in Table \ref{tab:Detail-generator}.

\begin{table}[!t]\small
\centering
\caption{The architecture of color-liked space generator. \textit{Conv} is the operation of convolutional layer and \textit{BN} is the operation of batch normalization. $w$ and $h$ are the size of input raw image.}\label{tab:Detail-generator}
\label{my-label}
\begin{tabular}{|C{1.8cm}|C{5.5cm}|}
\hline
Output Size               & Layer                              \\ \hline \hline
$w\times h\times 3$       & Input $f(x,y)$                     \\ \hline
$w\times h\times 64$      & Conv,lReLU                         \\ \hline
$w\times h\times 64$      & Residual:Conv,BN,lReLU,Conv,BN,Sum \\ \hline
$w\times h\times 64$      & Residual:Conv,BN,lReLU,Conv,BN,Sum \\ \hline
$w\times h\times 64$      & Residual:Conv,BN,lReLU,Conv,BN,Sum \\ \hline
$w\times h\times 64$      & Residual:Conv,BN,lReLU,Conv,BN,Sum \\ \hline
$w\times h\times 64$      & Residual:Conv,BN,lReLU,Conv,BN,Sum \\ \hline
$w\times h\times 64$      & Conv,BN,Sum                        \\ \hline
$w\times h\times 64$      & Conv,lReLU                         \\ \hline
$w\times h\times 3$       & Conv                               \\ \hline
\end{tabular}
\end{table}

\subsection{Feature Extractor}
A \textit{perceptual similarity measure} is proposed by Dosovitskiy and Brox \cite{dosovitskiy2016generating} as well as Johnson \textit{et al.} \cite{johnson2016perceptual}. Rather than computing triplet loss in generated color-liked space, our proposed method first maps the generated color-liked space into a feature space by a differentiable function $\emptyset$. After feature extraction, the generated $f^{'}(x,y)$ can be written as $\emptyset (f^{'}(x,y),W_{2})$, where $W_{2}$ represents the parameters of feature extractor. Compared to generated color-liked space, $\emptyset (f^{'}(x,y),W_{2})$ allows color-liked space generator outputs that may not match the same class image with pixel-wise accuracy but instead encourages the generator to produce images that have similar feature representations.

For the feature mapping $\emptyset$, we exploit the pre-trained implementation of the popular VGG-19 \cite{simonyan2014very} network, which consists of stacked convolutional layers coupled with pooling operations to gradually decrease the spatial dimension of the image and to extract higher-level features in higher layers. To acquire high-level features, we use the last fully connected layer to describe the image obtained in the generated color-liked space.

\subsection{Points-to-Center Triplet Loss}
\begin{figure*}[!t]
\centering
\subfigure[Randomly combining triplet training samples.]{\includegraphics[width=0.35\textwidth]{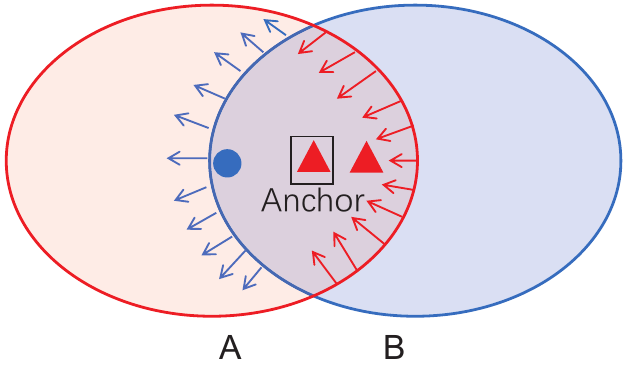}
\hfil
\label{fig:imperfection}}
\subfigure[Points-to-center combining mechanism.]{\includegraphics[width=0.35\textwidth]{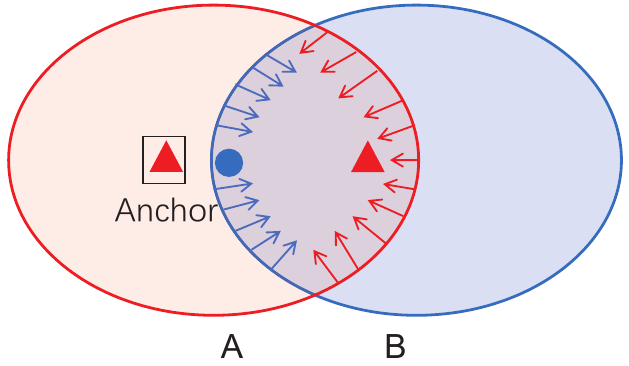}
\hfil
\label{fig:perfection}}
\caption{Two Different kinds of mechanisms for combining triplet training samples, where A and B represent two different classes.}
\label{fig:imperfectionANDperfection}
\end{figure*}

To begin  with, let $F^{+}=\{f^{+}_{1},f^{+}_{2},...,f^{+}_{i},...,f^{+}_{m}\}$ be the real face image set and $F^{-}=\{f^{-}_{1},f^{-}_{2},...,f^{-}_{j},...,f^{-}_{n}\}$ be the fake face image set, where $m$ and $n$ denote the number of different face images in $F^{+}$ and $F^{-}$, and $f^{+}_{i}$ and $f^{-}_{j}$ represent the $i_{th}$ and $j_{th}$ face image in $F^{+}$ and $F^{-}$, respectively. The goal of our color-liked space generator is to find a space mapping function that can minimize intraclass distance, maximize interclass distance and keep a safe margin between $F^{+}$ and $F^{-}$. In triplet network, conventional works randomly select triplet training samples to train the network \cite{Ding2015Deep,Wang2016Joint,Swami2016Triplet}. However, this kind of points to points selection mechanism restricts the stability of network training. As illustrated in Fig. \ref{fig:imperfection}, when randomly combining triplet training samples, the distribution of different classes may change in the same direction leading to the interclass distance cannot be maximized according to our goal.

In order to tackle the problem of triplet training sample combination, we chose a center image as the anchor image and combine triplet samples with it, as shown in Fig. \ref{fig:perfection}. More specifically, all face images in $F^{+}$ and $F^{-}$ are sequentially fed into the initialized color-liked space generator and pre-trained feature extractor, where the results are represented as $T^{+}=\emptyset(G(F^{+},W_{1}),W_{2})=\{t^{+}_{1},t^{+}_{2},...,t^{+}_{i},...,t^{+}_{m}\}$ and $T^{-}=\emptyset(G(F^{-},W_{1}),W_{2})=\{t^{-}_{1},t^{-}_{2},...,t^{-}_{i},...,t^{-}_{n}\}$. The center of real face features can be calculated as $\overline{t^{+}}=\frac{1}{m}\sum^{m}_{i=1} t^{+}_{i}$.

Then the anchor image of $F^{+}$ is the one that is a real face image with the nearest Euclidean distance from the feature center $\overline{t^{+}}$ and denoted as $f^{+}_{c}$, where $c$ is the index of the anchor image. Given a threshold $\tau$, if the Euclidean distance between the real face image and the anchor image is greater than $\tau$, the real face image and the anchor image form a image pair with same class represented as $\{f^{+}_{c},f^{+}_{p}\}$; if the Euclidean distance between the fake face image and the anchor image is less than $\tau$, then the fake face image and the anchor image compose an image pair with heterogeneous classes written as $\{f^{+}_{c},f^{-}_{p}\}$. Finally, the points to set triplet training samples are denoted as $\{f^{+}_{p},f^{+}_{c},f^{-}_{p}\}_{p=1}^{P}$, where $p$ and $P$ are the index and the number of triplet training samples, respectively. As aforementioned, our goal is to minimize the distance between $f^{+}_{p}$ and $f^{+}_{c}$, maximize the distance and keep a safe margin between $f^{-}_{p}$ and $f^{+}_{c}$. Therefore, the loss function of the goal can be formulated as Eq. \ref{Eq:tripletLoss}.
\begin{eqnarray}\label{Eq:tripletLoss}
\begin{split}
  \mathcal{L}=&\frac{1}{P}\sum^{P}_{p=1}max\{\\
  &||\emptyset(G(f^{+}_{c},W_{1}),W_{2})-\emptyset(G(f^{+}_{p},W_{1}),W_{2})||^{2}-\\
  &||\emptyset(G(f^{+}_{c},W_{1}),W_{2})-\emptyset(G(f^{-}_{p},W_{1}),W_{2})||^{2},\gamma\}
\end{split}
\end{eqnarray}
where $\gamma$ is the safe margin and the role of the max operation with $\gamma$ is to prevent the overall value of the loss function from being dominated by easily identifiable triplets \cite{Ding2015Deep}.

\subsection{Gradient of Parameters}
In training stage, the stochastic gradient descent (SGD) algorithm \cite{Bottou2010Large} is used to learn the network parameters. So, calculating the gradient of parameters is the first important thing. Let $d(f_{p},W_{1},W_{2})$ be the difference in distance between $\{f^{+}_{c},f^{+}_{p}\}$ and $\{f^{+}_{c},f^{-}_{p}\}$ in the triplet sample:
\begin{eqnarray}\label{Eq:Learning-d}
\begin{split}
  &d(f_{p},W_{1},W_{2})=\\
  &||\emptyset(G(f^{+}_{c},W_{1}),W_{2})-\emptyset(G(f^{+}_{p},W_{1}),W_{2})||^{2}-\\
                       &||\emptyset(G(f^{+}_{c},W_{1}),W_{2})-\emptyset(G(f^{-}_{p},W_{1}),W_{2})||^{2}
\end{split}
\end{eqnarray}
and the loss function can be rewritten as
\begin{eqnarray}\label{Eq:Learning-L}
  \mathcal{L}=\frac{1}{P}\sum^{P}_{p=1}max\{d(f_{p},W_{1},W_{2}),\gamma\}
\end{eqnarray}
Then the partial derivative of the loss function about $W_{1}$ can be calculated as Eq. \ref{Eq:Learning-partial-W1}.
\begin{eqnarray}\label{Eq:Learning-partial-W1}
  \frac{\partial\mathcal{L}}{\partial W_{1}}=\frac{1}{P}\sum^{P}_{p=1}h(f_{p})
\end{eqnarray}
\begin{eqnarray}\label{Eq:Learning-partial-h}
  h(f_{p})=
  \begin{cases}
  \frac{\partial d(f_{p},W_{1},W_{2})}{\partial W_{1}}    &if \ d(f_{p},W_{1},W_{2}) > \gamma \\
  0  &if \ d(f_{p},W_{1},W_{2}) \leq \gamma
  \end{cases}
\end{eqnarray}
Based on the definition of $d(f_{p},W_{1},W_{2})$, its gradient can be obtained as follows:
\begin{eqnarray}\label{Eq:Learning-partial-W1}
\begin{split}
  &\frac{\partial d(f_{p},W_{1},W_{2})}{\partial W_{1}}=\\
  &2(\emptyset(G(f^{+}_{c}))-\emptyset(G(f^{+}_{p})))\cdot \frac{\partial \emptyset(G(f^{+}_{c}))-\partial \emptyset(G(f^{+}_{p}))}{\partial W_{1}}-\\
  &2(\emptyset(G(f^{+}_{c}))-\emptyset(G(f^{-}_{p})))\cdot \frac{\partial \emptyset(G(f^{+}_{c}))-\partial \emptyset(G(f^{-}_{p}))}{\partial W_{1}}
\end{split}
\end{eqnarray}
where $\emptyset(G(f^{+}_{c}))$ is the abbreviation of $\emptyset(G(f^{+}_{c},W_{1}),W_{2})$, $\emptyset(G(f^{+}_{p}))$ is the abbreviation of $\emptyset(G(f^{+}_{p},W_{1}),W_{2})$, and $\emptyset(G(f^{-}_{p}))$ is the abbreviation of $\emptyset(G(f^{-}_{p},W_{1}),W_{2})$, respectively. For the parameter $W_{2}$ in feature extractor, its gradient can also be obtained as $W_{1}$.

\subsection{Implementation Details}
For the parameters in feature extractor, we transfer the pre-trained deep neural network (i.e. VGG-19) to initialize them. But for the parameters in color-liked space generator, they are initialized based on \cite{He2015Delving} as illustrated in Eq. \ref{Eq:initialization}, which can ensure that all convolutional layers in the network initially have the approximately same output distribution and empirically improve the rate of convergence.
\begin{eqnarray}\label{Eq:initialization}
  W_{1}=\frac{rand(n_l)}{\sqrt{2/n_l}}
\end{eqnarray}
where $rand(\cdot)$ samples from a zero mean, unit standard derivation gaussian function, and $n_l$ is the channel number of inputs in convolutional layer. In training stage, the momentum and weight decay of SGD are set to 0.9 and 0.0005, respectively. The learning rate is set to \(10^{-3}\) in the beginning and will be multiplied with damping factor \(0.5\) when all mini-batches are traversed and re-allocated randomly.

The inputs of face images are normalized into $224 \times 224 \times 3$ with the scale ranging from 0 to 1. In color-liked space generator, five residual blocks are concatenated to constitute the generator. For the threshold $\tau$, it is five times the average Euclidean distance between the real face images and the anchor image.
In testing stage, we extract the deep features of the learned face images from feature extractor and use an SVM classifier to predict whether the input face is a presentation attack. In our paper, we realize the proposed color-liked space generator and SVM based on the toolbox of MatConvNet with the version 1.0-beta20 \footnote{http://www.vlfeat.org/matconvnet/} and liblinear with the version 1.96 \footnote{https://www.csie.ntu.edu.tw/$\sim$cjlin/liblinear/} \cite{Fan2008LIBLINEAR}, respectively.

\section{Experimental Setup}

\subsection{Experimental Data}
Our proposed face PAD algorithm is validated on two publicly available face PAD databases: Replay-Attack \cite{Chingovska2012On} and OULU-NPU \cite{Boulkenafet2017OULU}. The detailed descriptions are given below.

\subsubsection{Replay-Attack}
\begin{figure}[!t]
 \centering
 \includegraphics[width=0.48\textwidth]{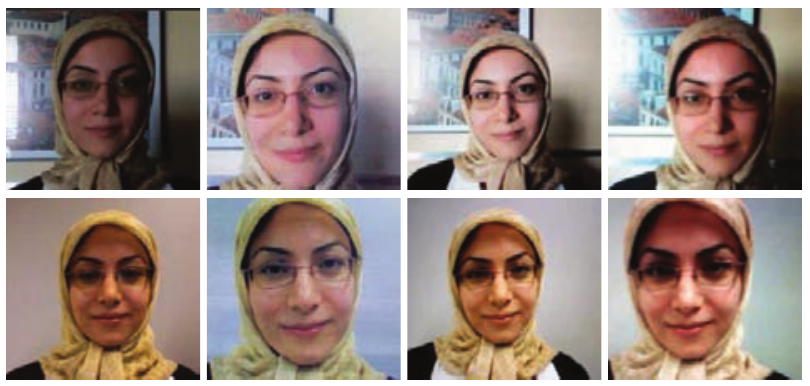}
 \caption{Samples from Replay-Attack database. The first row presents images taken from the controlled scenario,
while the second row corresponds to the images from the adverse scenario. From the left to the right: real faces and the corresponding printed photo, displayed image and replayed video attacks.}\label{replay_samples}
\end{figure}

The IDIAP Replay-Attack database \footnote{https://www.idiap.ch/dataset/replayattack/download-proc}  \cite{Chingovska2012On} consists of 1300 video clips of real and attack and attempts to 50 clients. These clients are divided into 3 subject-disjoint subsets for training, development and testing (15, 15 and 20, respectively). The real face videos are recorded under two different lighting conditions: \textit{controlled} and \textit{adverse}. Three types of attacks are created: printed photos, displayed images and replayed videos. In displayed image and replayed video attacks, high quality images and videos of real clients are replayed on iPhone 3GS and iPad display devices. For printed photo attacks, high quality images were printed on A4 papers and presented in front of the camera. Fig. \ref{replay_samples} shows some examples of real and fake faces.

\begin{figure}[!t]
 \centering
 \includegraphics[width=0.48\textwidth]{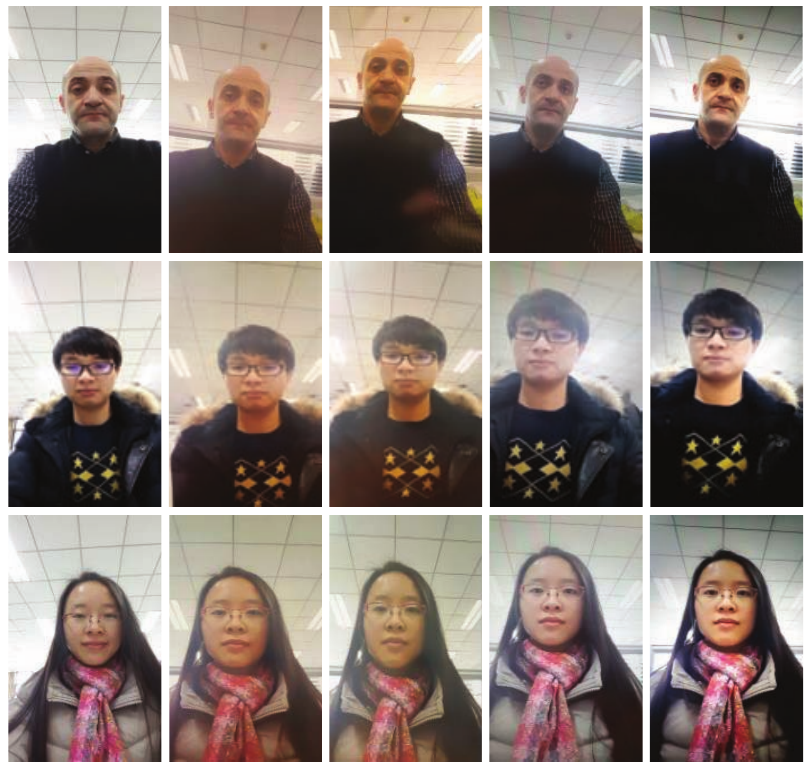}
 \caption{Samples from OULU-NPU database. From left to right: real faces and the corresponding printed photo, printed photo, replayed video and replayed video attacks.}\label{oulu_samples}
\end{figure}

\begin{table*}[!t]
\centering
\caption{The impact of different $\gamma$ on APER(\%), BPCER(\%) and ACER((\%) of Replay-Attack and OULU-NPU databases.}
\label{Tab:gama}
\begin{tabular}{|c|c|c|c|c|c|c|c|c|c|c|c|c|}
\hline
\multicolumn{13}{|c|}{Replay-Attack Database}                                                                                                                                                          \\ \hline \hline
\multirow{2}{*}{$\gamma$} & \multicolumn{3}{c|}{Overall attacks} & \multicolumn{3}{c|}{Printed photo} & \multicolumn{3}{c|}{Displayed image} & \multicolumn{3}{c|}{Replayed video} \\ \cline{2-13}
           & APCER     & BPCER     & ACER      & APCER       & BPCER       & ACER     & APCER       & BPCER       & ACER     & APCER       & BPCER       & ACER                          \\ \hline
0.1        &1.4        & 0.0       & 0.7       & 2.1         & 0.2         & 1.1      & 0.7         & 0.0         & 0.3      & 0.1         & 0.0         & 0.0         \\ \hline
0.5        &2.4        & 0.0       & 1.2       & 0.0         & 0.5         & 0.3      & 0.9         & 0.0         & 0.4      & 0.9         & 0.0         & 0.5         \\ \hline
1          &1.7        & 0.0       & 0.8       & 0.0         & 2.0         & 1.0      & 3.6         & 0.0         & 1.8      & 0.2         & 0.0         & 0.1         \\ \hline
5          &4.5        & 0.0       & 2.3       & 0.2         & 0.0         & 0.1      & 1.4         & 0.0         & 0.7      & 2.3         & 0.0         & 1.2         \\ \hline \hline
\multicolumn{13}{|c|}{OULU-NPU Database}                                                                                                                                                               \\ \hline \hline
\multirow{2}{*}{$\gamma$} & \multicolumn{3}{c|}{Overall attacks} & \multicolumn{3}{c|}{Printed photo} & \multicolumn{3}{c|}{Displayed image} & \multicolumn{3}{c|}{Replayed video} \\ \cline{2-13}
          & APCER      & BPCER     & ACER      & APCER       & BPCER       & ACER     & APCER       & BPCER       & ACER     & APCER       & BPCER       & ACER                              \\ \hline
0.1       & 8.7        & 10.5      & 9.6       & 4.2         & 6.2         & 5.2      & -           & -           & -        & 6.2         & 7.0         & 6.6         \\ \hline
0.5       & 3.3        & 9.4       & 6.3       & 4.9         & 3.7         & 4.3      & -           & -           & -        & 7.0         & 4.2         & 5.6         \\ \hline
1         & 6.0        & 7.0       & 6.5       & 3.1         & 6.6         & 4.9      & -           & -           & -        & 5.7         & 4.5         & 5.1         \\ \hline
5         & 9.7        & 9.9       & 9.8       & 1.3         & 12.6        & 7.0      & -           & -           & -        & 5.7         & 5.6         & 5.7         \\ \hline
\end{tabular}
\end{table*}

\subsubsection{OULU-NPU}
The OULU-NPU Database \footnote{https://sites.google.com/site/oulunpudatabase/welcome} \cite{Boulkenafet2017OULU} consists of 4950 real access and attack videos and attempts 55 clients. Similar to Replay-Attack database, all clients are divided into 3 subject-disjoint subsets for training, development and testing (20, 15 and 20, respectively). These videos were recorded using the front cameras of six mobile devices in three sessions with different illumination conditions and background scenes. Two types of fake faces are created: printed photo and replayed video attacks. The attacks were created using two printers and two display devices. For the replayed video attacks, the original face videos were recorded by 6 different cell phones. Fig. \ref{oulu_samples} shows some examples of real and fake faces captured from the first scenario.

\subsection{Evaluation Protocol}
For performance evaluation, the results are reported in term of recently standardized ISO/IEC 30107-3 metrics \cite{ISO2016}: Attack Presentation Classification Error Rate (APCER) and Bona Fide Presentation Classification Error Rate (BPCER). In principle, these two metrics correspond to the FAR and FRR commonly used in the PAD related literature. However, different with the (false acceptance rate) FAR and false rejection rate (FRR), the attacker's potential (such as expertise, resources and motivation) in the ��worst case scenario�� are taken into considered by APCER and BPCER. It is noted that the APCER and BPCER depend on the decision threshold. Therefore, the development set is used to fine tuning the system parameters and estimate the threshold value. To compare the overall system performance in a single value, the Average Classification Error Rate (ACER) is computed which is the average of the APCER and the BPCER at the decision threshold estimated by Equal Error Rate (EER) on the development set.

\section{Experimental Results and Discussion}
In this section, we present and discuss the detection results obtained in the generated color-liked space. Firstly, the impact of hyper-parameters $\gamma$ is explored. Then, we compare our proposed points to center triplet training sample combination mechanism with conventional randomly combination mechanism. After that, we visualize the distribution of real and fake faces in the proposed color-liked space and compare the detection results with that computed in existing color spaces. Finally, the performance of our method is compared against the state-of-the-art algorithms and the generalization capabilities of learned color-liked space are evaluated by conducting cross-database experiments.

\subsection{Safe Margin for Feature Extractor}
In this part, the influence of the hyper-parameter $\gamma$ on detection performance is presented and its optimal value is also been determined. Table \ref{Tab:gama} illustrates the detection results of different $\gamma$. It can be clearly seen that $\gamma$ has a great influence on the performance. For example, when it is set to 0.1, the ACER of replayed video attack of Replay-Attack database is 0.0\%. However, when it is set to 5, the ACER changes to 1.2\%. For how to set the value of $\gamma$, we find that the optimal values for different databases are different. More specifically, for Replay-Attack database, $\gamma$ should be set to 0.1 with the averaged ACER=0.5\%; but for OULU-NPU database, $\gamma$ should be set to 1 with the averaged ACER=5.4\%. While considering the averaged ACER of both Replay-Attack and OULU-NPU databases, we set the $\gamma$ to 0.5 in our proposed method. In addition, the detection performance does not improve as the increase of safe margin between the real faces and fake faces. We conjecture the reason may lie in that larger $\gamma$ will cause the network to over-fit during training. Especially for face PAD, the training data is very limited.

\begin{table*}[!t]
\centering
\caption{The Comparison between our points to center (P2C) sample combination mechanism and randomly combination (RC) mechanism.}
\label{Tab:P2S}
\begin{tabular}{|c|c|c|c|c|c|c|c|c|c|c|c|c|}
\hline
\multicolumn{13}{|c|}{Replay-Attack Database}                                                                                                                                                          \\ \hline \hline
\multirow{2}{*}{Mechanism} & \multicolumn{3}{c|}{Overall attacks} & \multicolumn{3}{c|}{Printed photo} & \multicolumn{3}{c|}{Displayed image} & \multicolumn{3}{c|}{Replayed video} \\ \cline{2-13}
          & APCER   & BPCER   & ACER   & APCER   & BPCER    & ACER   & APCER    & BPCER    & ACER    & APCER   & BPCER    & ACER            \\ \hline
RC        & 4.0     & 0.0     & 2.1    & 2.7     & 0.0      & 1.4    & 0.8      & 0.0      & 0.4     & 0.4     & 0.0      & 0.2             \\ \hline
P2C       & 2.4     & 0.0     & 1.2    & 0.0     & 0.5      & 0.3    & 0.9      & 0.0      & 0.4     & 0.9     & 0.0      & 0.5             \\ \hline \hline
\multicolumn{13}{|c|}{OULU-NPU Database}                                                                                                                                                               \\ \hline \hline
\multirow{2}{*}{Mechanism} & \multicolumn{3}{c|}{Overall attacks} & \multicolumn{3}{c|}{Printed photo} & \multicolumn{3}{c|}{Displayed image} & \multicolumn{3}{c|}{Replayed video} \\ \cline{2-13}
          & APCER   & BPCER   & ACER   & APCER   & BPCER    & ACER   & APCER    & BPCER    & ACER    & APCER   & BPCER    & ACER            \\ \hline
RC        & 10.5    & 17.1    & 13.8   & 7.3     & 7.3      & 7.3    & -        & -        & -       & 4.7     & 7.8      & 6.2             \\ \hline
P2C       & 3.3     & 9.4     & 6.3    & 4.9     & 3.7      & 4.3    & -        & -        & -       & 7.0     & 4.2      & 5.6             \\ \hline
\end{tabular}
\end{table*}

\subsection{Points-to-Center Combination Mechanism}
As aforementioned, the triplet training samples are combined based on the proposed points-to-center (P2C) mechanism instead of randomly combination (RC). Table \ref{Tab:P2S} shows the detection results obtained based on these two different combination mechanisms. From the table we can find that our proposed P2C mechanism significantly improves overall system performance, except for the ACER of replayed video attack of Replayed-Attack database. Comparing the detection results of the two databases, one conclusion can be obtained that the performance improvement of P2C mechanism is more obvious for the OULU-NPU database. More specifically, the ACER of overall OULU-NPU database has dropped more than 54\%. But for the overall Replayed-Attack database, the ACER has dropped by 43\%. The reason may lie in that the distribution of real and fake faces in Replay-Attack database is simpler than that of OULU-NPU database, which limits the superiority of P2C.

\begin{table*}[!t]
\centering
\caption{Comparison between our generated color-liked space and existing color spaces.}
\label{Tab:ComparisonWithExistingSpace}
\begin{tabular}{|c|c|c|c|c|c|c|c|c|c|c|c|c|}
\hline
\multicolumn{13}{|c|}{Replay-Attack Database}                                                                                                                                                          \\ \hline \hline
\multirow{2}{*}{Space} & \multicolumn{3}{c|}{Overall attacks} & \multicolumn{3}{c|}{Printed photo} & \multicolumn{3}{c|}{Displayed image} & \multicolumn{3}{c|}{Replayed video} \\ \cline{2-13}
            & APCER     & BPCER     & ACER      & APCER       & BPCER       & ACER     & APCER       & BPCER       & ACER     & APCER       & BPCER       & ACER                          \\ \hline
RGB         &8.4        & 2.7       & 5.6       & 4.2         & 6.9         & 5.5      & 7.5         & 2.9         & 5.2      & 4.2         & 0.0         & 2.1     \\ \hline
HSV         &14.9       & 9.0       & 11.9      & 2.5         & 5.2         & 3.8      & 15.0        & 9.5         & 12.2     & 11.7        & 4.5         & 8.1     \\ \hline
YCbCr       &11.3       & 10.5      & 10.9      & 2.8         & 3.4         & 3.1      & 11.0        & 5.5         & 8.2      & 5.5         & 4.9         & 5.2     \\ \hline
Our Method  &2.4        & 0.0       & 1.2       & 0.0         & 0.5         & 0.3      & 0.9         & 0.0         & 0.4      & 0.9         & 0.0         & 0.5     \\ \hline \hline
\multicolumn{13}{|c|}{OULU-NPU Database}                                                                                                                                                               \\ \hline \hline
\multirow{2}{*}{Space} & \multicolumn{3}{c|}{Overall attacks} & \multicolumn{3}{c|}{Printed photo} & \multicolumn{3}{c|}{Displayed image} & \multicolumn{3}{c|}{Replayed video} \\ \cline{2-13}
            & APCER      & BPCER     & ACER      & APCER       & BPCER       & ACER     & APCER       & BPCER       & ACER     & APCER       & BPCER       & ACER                              \\ \hline
RGB         & 18.7       & 11.9      & 15.3      & 14.6        & 9.9         & 12.3     & -           & -           & -        & 19.9        & 13.1        & 16.5       \\ \hline
HSV         & 14.9       & 19.0      & 17.0      & 7.3         & 10.6        & 8.9      & -           & -           & -        & 14.6        & 16.8        & 15.7       \\ \hline
YCbCr       & 17.8       & 9.3       & 13.6      & 13.0        & 11.9        & 12.5     & -           & -           & -        & 10.3        & 6.7         & 8.5        \\ \hline
Our Method  & 3.3        & 9.4       & 6.3       & 4.9         & 3.7         & 4.3      & -           & -           & -        & 7.0         & 4.2         & 5.6        \\ \hline
\end{tabular}
\end{table*}

\begin{table*}[!t]
\centering
\caption{Comparison between our proposed countermeasure and state-of-the-art methods on the two benchmark databases using the based evaluation.}
\label{Tabel:Comparison}
\begin{tabular}{|C{2.9cm}|C{0.9cm}|C{0.9cm}|C{0.9cm}|C{0.9cm}|C{0.9cm}|C{0.9cm}|C{0.9cm}|C{0.9cm}|C{0.9cm}|C{0.9cm}|}
\hline
\multirow{2}{*}{Methods} & \multicolumn{5}{c|}{Replay-Attack} & \multicolumn{5}{c|}{OULU-NPU} \\ \cline{2-11}
                                                & EER         & HTER        & APCER      & BPCER      & ACER       &EER          & HTER        & APCER     &BPCER    &ACER  \\ \hline \hline
LBP-TOP \cite{Pereira2012LBP-TOP}               & 7.9         & 7.6         & -          & -          & -          & -           & -           & -         & -       & - \\ \hline
LDP-TOP \cite{Phan2016FACE}                     & 2.5         & 1.7         & -          & -          & -          & -           & -           & -         & -       & - \\ \hline
DMD \cite{Tirunagari2015Detection}              & 5.3         & 3.7         & -          & -          & -          & -           & -           & -         & -       & - \\ \hline
Scale LBP \cite{Boulkenafet2016Scale}           & 0.7         & 3.1         & -          & -          & -          & -           & -           & -         & -       & - \\ \hline
Deep CNN \cite{Yang2014Learn}                   & 6.1         & 2.1         & -          & -          & -          & -           & -           & -         & -       & - \\ \hline
3D CNN \cite{Li2018Learning}                    & \textbf{0.3}& 1.2         & -          & -          & -          & -           & -           & -         & -       & - \\ \hline
Image Quality \cite{Galbally2014Image}$\dag$    & -           & 15.2        & 17.9+      & 12.5       & 15.2+      & 31.9        & 36.3        & 56.4      & 24.5    & 40.5 \\ \hline
RGB LBP \cite{Boulkenafet2016FaceTIFS}$\ddag$   & 7.3         & 11.2        & 10.1       & 14.3       & 12.2       & 11.2        & 18.5        & 18.8      & 20.0    & 19.4  \\ \hline
HSV LBP \cite{Boulkenafet2016FaceTIFS}$\ddag$   & 7.7         & 9.8         & 18.5       & 6.1        & 13.7       & 12.0        & 15.5        & 21.1      & 15.9    & 18.5  \\ \hline
YCbCr LBP \cite{Boulkenafet2016FaceTIFS}$\ddag$ & 4.1         & 7.4         & 14.6       & 8.9        & 10.4       & 14.2        & 17.8        & 21.7      & 13.0    & 17.3  \\ \hline
\hline
Color-liked Space LBP                           & 2.8         & 2.9         & 5.0        & 4.6        & 4.8        & 8.6         & 12.2        & 14.9      & 10.5    & 12.7 \\ \hline
Color-liked Space VGG                           & 0.8         & \textbf{0.7}&\textbf{2.4}&\textbf{0.0}&\textbf{1.2}& \textbf{7.6}& \textbf{6.0}&\textbf{3.3}&\textbf{9.4}&\textbf{6.3} \\ \hline
\end{tabular}
\begin{tablenotes}
    \item $+$ means the actual value is greater than the value.
    \item $\dag$ was retested on OULU-NPU database.
    \item $\ddag$ was retested on Replay-Attack and OULU-NPU databases.
\end{tablenotes}
\end{table*}

\begin{figure*}[!t]
\centering
\subfigure[Replay-Attack database in the generated color-liked space.]{\includegraphics[width=0.8\textwidth]{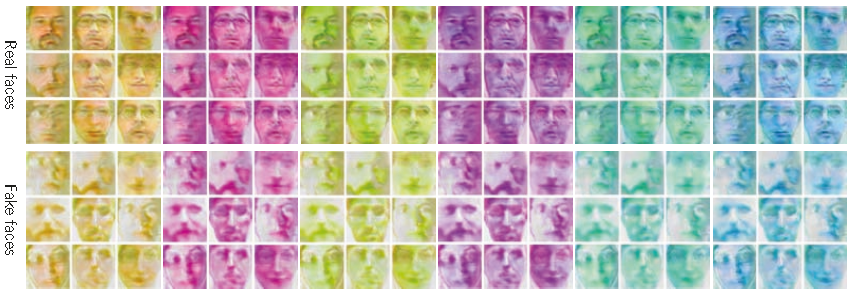}
\hfil
\label{fig:Replay-real-fake}}
\subfigure[OULU-NPU database in the generated color-liked space.]{\includegraphics[width=0.8\textwidth]{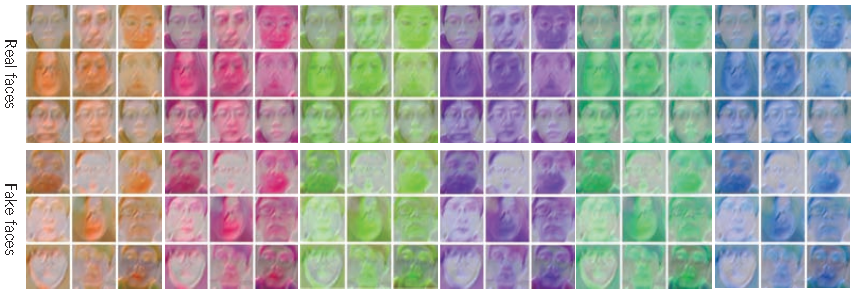}
\hfil
\label{fig:OULU-real-fake}}
\caption{Visualize some randomly selected real and fake face images of Replay-Attack and OULU-NPU databases in the generated color-liked space. From left to right, different colors are obtained by different color-liked channels combination.}
\label{fig:visualize-color-liked-space}
\end{figure*}

\begin{figure}[!h]
 \centering
 \includegraphics[width=0.3\textwidth]{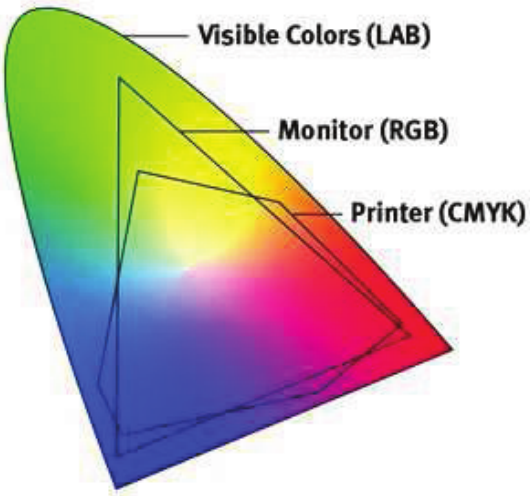}
 \caption{Colour gamut of a monitor (RGB) and a printer (CMYK) compared with the whole pallet of visible colours \cite{Boulkenafet2016FaceTIFS}.}\label{fig:color-gamut}
\end{figure}

\begin{figure*}[!t]
\centering
\subfigure[RGB color space.]{\includegraphics[width=0.29\textwidth]{RGB_distribution.pdf}
\hfil
\label{fig:Replay-RGB}}
\subfigure[Overall attacks in color-liked space.]{\includegraphics[width=0.29\textwidth]{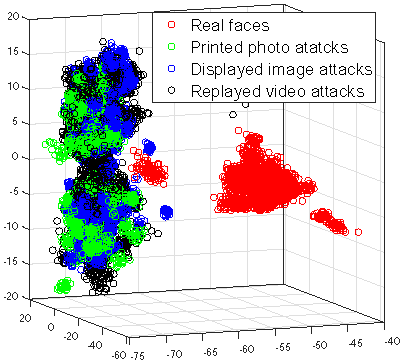}
\hfil
\label{fig:Replay-overall}}
\subfigure[Printed photo attacks in color-liked space.]{\includegraphics[width=0.29\textwidth]{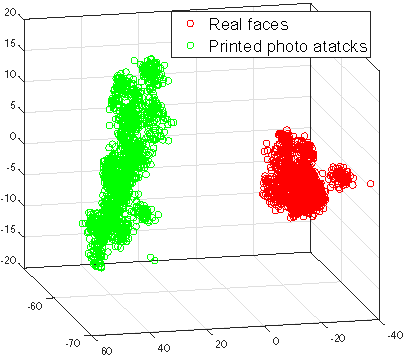}
\hfil
\label{fig:Replay-printed}}
\subfigure[Displayed image attacks in color-liked space.]{\includegraphics[width=0.29\textwidth]{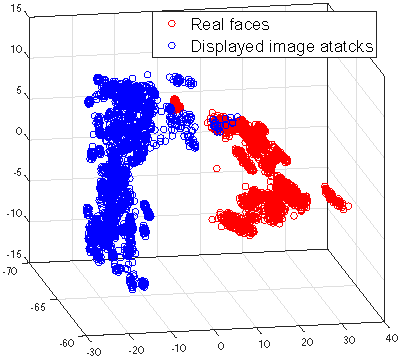}
\hfil
\label{fig:Replay-displayed}}
\subfigure[Replayed video attacks in color-liked space.]{\includegraphics[width=0.29\textwidth]{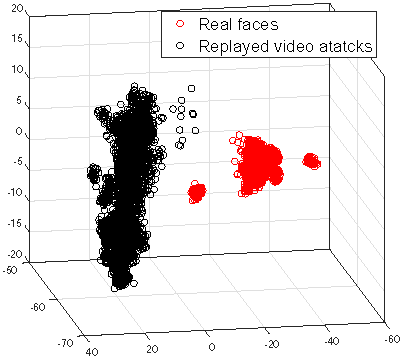}
\hfil
\label{fig:Replay-replayed}}
\caption{The distribution of Replay-Attack database in RGB color space and our generated color-liked space. For the figures (a) to (e), we extract deep features by the feature extractor and visualize them in a 3D space.}
\label{fig:Replay-Attack-Distribution-Compare}
\end{figure*}

\begin{figure*}[!t]
\centering
\subfigure[RGB color space.]{\includegraphics[width=0.3\textwidth]{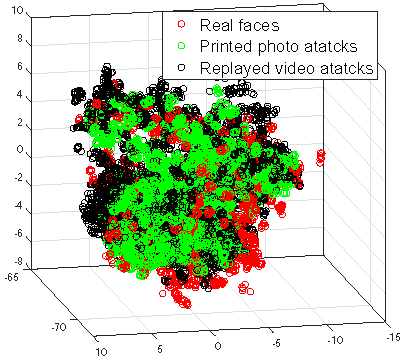}
\hfil
\label{fig:OULU-RGB}}
\subfigure[Overall attacks in color-liked space.]{\includegraphics[width=0.3\textwidth]{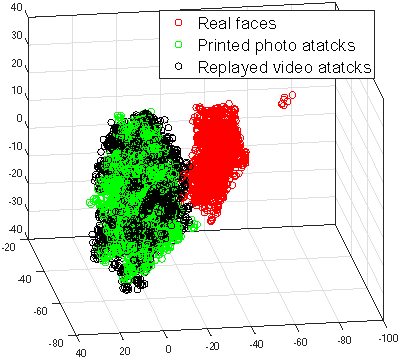}
\hfil
\label{fig:OULU-overall}} \\
\subfigure[Printed photo attacks in color-liked space.]{\includegraphics[width=0.3\textwidth]{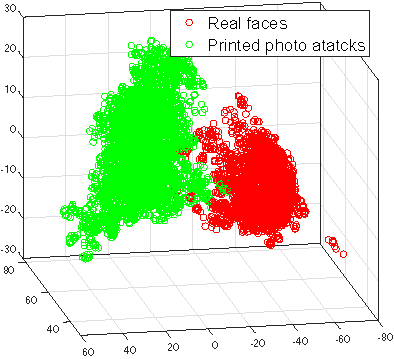}
\hfil
\label{fig:OULU-printed}}
\subfigure[Replayed video attacks in color-liked space.]{\includegraphics[width=0.3\textwidth]{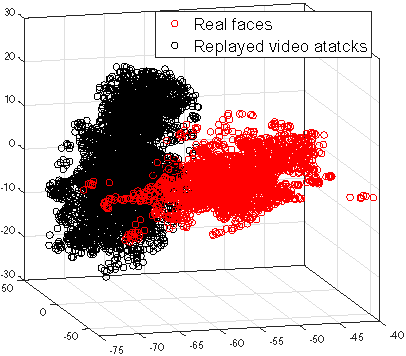}
\hfil
\label{fig:OULU-replayed}}
\caption{The distribution of OULU-NPU database in RGB color space and our generated color-liked space. For the figures (a) to (d), we extract deep features by the feature extractor and visualize them in a 3D space.}
\label{fig:OULU-NPU-Distribution-Compare}
\end{figure*}

\subsection{Visualize the Data Distribution and Color-liked Space}
For a clearer analysis of the generated color-liked space, the sample distributions of Replay-Attack and OULU-NPU databases are described in Fig. \ref{fig:Replay-Attack-Distribution-Compare} and Fig. \ref{fig:OULU-NPU-Distribution-Compare}, respectively. From the figures \ref{fig:Replay-RGB} and \ref{fig:OULU-RGB}, we can clearly see that the real and fake face images are overlapped in RGB color space. However, in our generated space, the real and fake faces can be well separated, and there is a safe margin between them. By comparing Fig. \ref{fig:Replay-Attack-Distribution-Compare} and Fig. \ref{fig:OULU-NPU-Distribution-Compare}, we can find that the Replay-Attack database is easier to be addressed than the OULU-NPU database, which is consistent with the results shown in Table \ref{Tab:gama}.

Apart from plotting the distribution, the generated color-liked space is also been visualized in Fig. \ref{fig:visualize-color-liked-space}. Since the output of the generator has three channels, we visualize the randomly selected real and the fake faces in the learned color-liked space according to different channel combination sequences. From the visualization results, the difference between the real and fake faces is mainly revealed in color saturation. More specifically, the real faces are more saturated than the fake faces, which can be explained by the display colour gamut differences shown in Fig. \ref{fig:color-gamut}. Compared Fig. \ref{fig:visualize-color-liked-space} with Fig. \ref{fig:color-gamut}, we can summarize that the generated color-liked space learning process essentially magnifies the differences between the color saturation of real and fake faces.

\subsection{Comparison with Existing Color Spaces}
Furthermore, we extract the last full connected layer of VGG-19 from the learned color-liked space and existing color spaces (i.e. RGB, HSV and YCbCr), and compare the detection results in Table \ref{Tab:ComparisonWithExistingSpace}. Based on the comparison, we can observe that our proposed color-liked space has greatly improved the detection performance. For instance, in RGB color space, the averaged APCER, BPCER and ACER of Replay-Attack and OULU-NPU are 6.1\%, 3.1\%, 4.6\% and 17.7\%, 11.6\% 14.7\%, respectively. However, the averaged protocols obtained in our color-liked space are 1.1\%, 0.1\%, 0.6\% and 5.1\%, 5.8\%, 5.4\%. This indicates that the color-liked space makes it easier to distinguish the real and fake faces compared to RGB, HSV or YCbCr spaces.

\subsection{Comparison with the State of the Art}
In order to be consistent with the original evaluation protocol and to facilitate the later works compared with ours, we show the original EER, HTER \cite{Chingovska2012On} and the latest APCER, BPCER, ACER. Table \ref{Tabel:Comparison} gives a comparison with the state-of-the-art face PAD approaches proposed in the literature. Apart from extracting VGG features, we also extract LBP features from the learned color-liked space. It can be seen that our proposed color-liked space based method outperforms the-state-of-the-art algorithms on Replay-Attack database. Especially compared with \cite{Li2018Learning}, the HTER of VGG features is significantly reduced to 0.7\%, even though the EER is not as good as \cite{Li2018Learning}. For hand-crafted feature, the LBP extracted from our learned color-liked space takes on better superiority compared with \cite{Boulkenafet2016FaceTIFS} that extracts LBP features from existing color spaces. Regarding the OULU-NPU database, we compare the proposed method with its baseline and \cite{Galbally2014Image}. From the results, our learned color-liked space based method also significantly surpasses them. As for the learned color-liked space, we find that the extracted VGG features can get better detection results than LBP features. This can be interpreted as the generated color-liked space is learned based on VGG features rather than LBP.

\begin{table*}[!t]
\centering
\caption{The performance of cross-database experiment on the Replay-Attack and OULU-NPU databases compared with baseline method.}
\label{label:crossResults}
\begin{tabular}{|C{3cm}|C{1.6cm}|C{1.6cm}|C{1.6cm}|C{1.6cm}|C{1.6cm}|C{1.6cm}|}
\hline
\multirow{2}{*}{Methods} & \multicolumn{3}{c|}{\begin{tabular}[c]{@{}c@{}}Train on: Replay-Attack Evaluate on: OULU-NPU\end{tabular}} & \multicolumn{3}{c|}{\begin{tabular}[c]{@{}c@{}}Train on: OULU-NPU Evaluate on: Replay-Attack\end{tabular}} \\ \cline{2-7}
                                         & Train set     & Dev set       & Test set      & Train set     & Dev set         & Test set      \\ \hline \hline
Image Quality \cite{Galbally2014Image}   & 50.4          & 50.1          & 48.8          & 52.4          & \textbf{36.8}   & 39.1          \\ \hline
RGB LBP\cite{Boulkenafet2016FaceTIFS}    & 53.0          & 53.4          & 52.7          & 56.9          & 56.0            & 56.1          \\ \hline
HSV LBP\cite{Boulkenafet2016FaceTIFS}    & 51.1          & 52.8          & 51.6          & 51.7          & 52.9            & 50.0          \\ \hline
YCbCr LBP\cite{Boulkenafet2016FaceTIFS}  & 48.4          & 52.1          & 52.7          & 47.4          & 43.5            & 44.8          \\ \hline \hline
Color-liked Space LBP                    & 49.9          & 47.1          & \textbf{44.2} & 46.7          & 48.8            & 48.3          \\ \hline
Color-liked Space VGG                    & \textbf{44.7} & \textbf{44.4} & 46.2          & \textbf{41.6} & 42.6            & \textbf{36.2} \\ \hline
\end{tabular}
\end{table*}

\subsection{Cross-Database Analysis}
In real-world applications, face PAD techniques are operated in open environments, where the conditions and attack scenario are unknown. The conduct cross-database evaluation is conducted to gain insight into the generalization capabilities of our learned color-liked space. More specifically, the color-liked space generator is trained and tuned on one of the databases and then tested on another database. The computed ACERs are summarized in Table \ref{label:crossResults}.

When the generator is trained on OULU-NPU and tested on Replay-Attack, we notice that the averaged ACER of VGG features is 40.1\%. When the generator is trained on Replay-Attack and tested on OULU-NPU, the averaged  metric is 45.1\%. From these results, we conclude that the generator trained on Replay-Attack is not able to be generalized as good as trained on OULU-NPU. It is caused that the OULU-NPU database contains more variations in the collecting environment (e.g., light and camera quality) compared to Replay-Attack. Finally, compared with the baseline of color LBP based method \cite{Boulkenafet2016FaceTIFS}, our proposed method is more stable.

Although our proposed method can work well in intra database and cross database tests, it is still difficult to deal with face PAD when the light of collecting environment is variant. This is due to the fact that the brightness of captured face images can affect the distribution of colour gamut. In addition to light variation, the quality of the camera is also an important factor in the effectiveness of our detection method. Especially for the camera is not good enough to capture color information, our proposed countermeasure will may not be applicable.

\section{Conclusion}
In this paper, we addressed the problem of face PAD from the viewpoint of the color space analysis. Based on triplet training and perceptual similarity measure mechanisms, a new color-liked space was generated for fake face detection. Instead of randomly combining triplet training samples, we proposed a points-to-center combination method to solve the problem of instability in training costs. Extensive experiments on two latest and challenging presentation attack databases (the Replay-Attack and OULU-NPU) showed excellent results. On OULU-NPU database, the proposed color-liked space based method outperformed the baseline, while very competitive results were achieved on Replay-Attack database. Furthermore, in our cross-database evaluation, our proposed method showed promising generalization capabilities. Overall, from the results of Replay-Attack and OULU-NPU databases, we find that external-environment factors (e.g. light and camera quality) limit the effectiveness of our proposed detection method. Thus, we will focus on how to eliminate the influence of external environmental factors and improve the robustness of our method. Finally, the proposed method just chosen one anchor image from the real face images, which means the proposed method only minimized the intraclass distance in real face set. Therefore, we will also try to choose the second anchor image from the fake face images and minimize the intraclass distance in real and fake face sets.


\ifCLASSOPTIONcaptionsoff
  \newpage
\fi




\bibliography{spoofingNew2}

\vfill
\end{document}